# An Enhanced Adaptive Bi-clustering Algorithm through Building a Shielding Complex Sub-Matrix

Kaijie Xu [a]

[a] School of Electronic Engineering, Xidian University, Xi'an 710071, China

**ABSTRACT** Bi-clustering refers to the task of finding sub-matrices (indexed by a group of columns and a group of rows) within a matrix of data such that the elements of each sub-matrix (data and features) are related in a particular way, for instance, that they are similar with respect to some metric. In this paper, after analyzing the well-known Cheng and Church (CC) bi-clustering algorithm which has been proved to be an effective tool for mining co-expressed genes. However, Cheng and Church bi-clustering algorithm and summarizing its limitations (such as interference of random numbers in the greedy strategy; ignoring overlapping bi-clusters), we propose a novel enhancement of the adaptive bi-clustering algorithm, where a shielding complex sub-matrix is constructed to shield the bi-clusters that have been obtained and to discover the overlapping bi-clusters. In the shielding complex sub-matrix, the imaginary and the real parts are used to shield and extend the new bi-clusters, respectively, and to form a series of optimal bi-clusters. To assure that the obtained bi-clusters have no effect on the bi-clusters already produced, a unit impulse signal is introduced to adaptively detect and shield the constructed bi-clusters. Meanwhile, to effectively shield the null data (zero-size data), another unit impulse signal is set for adaptive detecting and shielding. In addition, we add a shielding factor to adjust the mean squared residue score of the rows (or columns), which contains the shielded data of the sub-matrix, to decide whether to retain them or not. We offer a thorough analysis of the developed scheme. The experimental results are in agreement with the theoretical analysis. The results obtained on a publicly available real microarray dataset show the enhancement of the bi-clusters performance thanks to the proposed method.

## I. INTRODUCTION

The traditional clustering algorithms analyze only the properties of the data samples (the number and types of the attributes or variables), but do not focus on the components of data such as Data-table, Data-column, Data-relation etc. This is a major issue affecting the clustering performance [1], especially when dealing with high-dimensional genes expression data, which motivates the development of the bi-clustering algorithm. Bi-clustering is not only able to reveal the global structure (as the traditional methods do) in data, but also able to discover the local information (it can discover clusters in the feature space and the data space simultaneously). In addition, Bi-clustering technology is also considered to be an effective tool for dealing with the high-dimensional data. Bi-clustering, after its birth, has received much attention and become one of the focal points in the data mining community. Bi-clustering was first introduced by Hartigan [2], and has been further developed since Cheng and Church proposed a bi-clustering algorithm based on variance and applied it to gene expression data [3]. Their work remains the most important contribution to the bi-clustering field.

At present, bi-clustering is the most widely used technology in the field of bioinformatics. Unlike traditional clustering methods that treat similarity (distance-based measures) as a function of pairs of genes or pairs of conditions, which is not applicable in high-dimensional space, the bi-clustering model measures coherence within the subset of genes and conditions. This model may be particularly useful in disclosing the involvement of genes or conditions in multiple pathways, some of which can only be discovered under the dominance of more consistent ones [4]. The coherence score [3] is defined as a symmetric function of genes and conditions involved, and therefore the bi-clustering is a process of simultaneous grouping of genes and conditions. The so-called mean squared residue (MSR) [3] is employed and applied to expression data transformed by a logarithm and augmented by the additive inverse. Bi-clustering is also referred in the literature as co-clustering and direct clustering, among other names, and has also been used in fields such as information retrieval and data mining.

Popular bi-clustering algorithms, such as Cheng and Church (CC) algorithm, FLOC [5], Plaid [6], OPSM [7], ISA [8], Spectral [9], xMOTIFs [10], and BiMax [11] have drawn much attention in the literature. Newer algorithms, such as Bayesian Bi-clustering [12], COALESCE [13], CPB [14], QUBIC [15], and FABIA [16] have not been extensively studied. Among them, the CC algorithm is the earliest and most studied one, and the newer algorithms are mostly based on the idea of the CC algorithm.

So far, many bi-clustering algorithms have been proposed; however, as of now the research on the bi-clustering is still at its initial stage. For the enrichment and development of the bi-clustering algorithms, in this paper, we first present a brief analysis of the well-known CC algorithm and elaborate on some related bi-clustering concepts. Meanwhile we discuss the advantages and drawbacks of the CC algorithm. In reference to the drawbacks of the CC algorithm (such as interference of random numbers in the greedy strategy; ignoring overlapping bi-clusters), we design an improved adaptive bi-clustering algorithm by building a shielding complex sub-matrix to adaptively shield the obtained bi-clusters and to discover new ones.

In the implementation of the new bi-clustering algorithm, we add an imaginary part to the constructed bi-clusters to increase their MSR and also set a shielding factor to adjust the MSR increments of the row (or column) which contains elements in the constructed bi-clusters to effectively shield the constructed bi-clusters. In addition, we set two unit impulse signals to adaptively detect the constructed bi-clusters and the null data (zero-size data) of the dataset to avoid the over-shielding and the shielding failure, so as to adaptively improve the bi-clusters. A detailed analysis and a comprehensive suite of experiments are provided. The experimental studies demonstrate that the proposed approach achieves better performance compared with that of the well-known CC algorithm. To the best of our knowledge, the idea of the proposed approach has not been considered in the previous studies.

This paper is organized as follows. The CC algorithm and bi-clustering related ideas are briefly reviewed in Section II. A novel enhancement of the adaptive bi-clustering algorithm is detailed in Section III. Section IV includes experimental setup and covers an analysis of completed experiments. Section V contains some conclusions.

## II. BI-CLUSTERING: DEFINITIONS AND PROBLEM FORMULATION

Consider a sample-feature expression matrix $A = (a_{ij})_{n \times m}$, where there are $n$ rows representing $n$ samples (data), $m$ columns representing $m$ features, and the entry $a_{ij}$ denotes the expression level of feature $j$ in sample $i$. Let $S = \{S_1, S_2, \cdots S_i, \cdots, S_n\}$ be the sample set, where $S_i = \{a_{i1}, a_{i2}, \cdots, a_{im}\}$ is called the feature vector of sample $i$. Similarly, for the features, it is denoted by $F = \{F_1, F_2, \cdots, F_j, \cdots, F_m\}$ with each vector $F_j = \{a_{1j}, a_{2j}, \cdots, a_{nj}\}^T$ being a column vector. Thus, we have $A = (S_1, S_2, \cdots, S_n)^T = (F_1, F_2, \cdots, F_m)$. A bi-cluster is a sub-matrix of data matrix, denoted by $B_k = (S_k, F_k)$ satisfying that $S_k \subseteq S$, $F_k \subseteq F$ and an entry denotes an intersection entry with corresponding row (sample) and column in both $A$ and $B_k$. Assume that there are $K$ bi-clusters found in data matrix $A$; the set of bi-clusters is denoted by $B = \{B_k : k = 1, 2, \cdots, K\}$. Usually, we use $(S_k, F)$ to denote a cluster of rows (samples) and $(S, F_k)$ a cluster of columns (features). Additionally, $|S_k|$ denotes the cardinality of $S_k$, i.e., the number of samples in bi-cluster $B_k = (S_k, F_k)$ while $|F_k|$ denotes the number of features. Clearly, we have $|S| = n$ and $|F| = m$.

Given a data matrix $A$, the bi-clustering problem is to design algorithms to find bi-clusters $B = \{B_k : k = 1, 2, \cdots, K\}$ of it, i.e., a sub-set of matrices of $A$ such that samples (rows $S_k$) of each bi-cluster $B_k$ exhibit some similar behavior under the corresponding features (columns, $F_k$). In other words, the bi-clustering problem is to identify a set of bi-clusters $B_k = (S_k, F_k)$ such that each bi-cluster $B_k$ satisfies some specific characteristics of homogeneity [17].

For a bi-cluster $B_k = (S_k, F_k)$, several means based on the bi-cluster are defined. The mean of row $i$ of $B_k$ is [18]

$$\mu_{ik}^{(r)} = \frac{1}{|F_k|} \sum_{j \in F_k} a_{ij} \tag{1}$$

the mean of column $j$ of $B_k$ is

$$\mu_{jk}^{(c)} = \frac{1}{|S_k|} \sum_{i \in S_k} a_{ij} \tag{2}$$

and the mean of all the entries in $B_k$ is

$$\mu_k = \frac{\sum_{i \in S_k} \sum_{j \in F_k} a_{ij}}{|S_k||F_k|} \tag{3}$$

The residue [3] of the entry $a_{ij}$ in bi-cluster $B_k$ is

$$r_{ij} = a_{ij} - \mu_{ik}^{(r)} - \mu_{jk}^{(c)} + \mu_k \tag{4}$$

the variance of bi-cluster $B_k$ is

$$\mathrm{Var}(B_k) = \sum_{i \in S_k} \sum_{j \in F_k} \left( a_{ij} - \mu_k \right)^2 \tag{5}$$

and mean squared residue (MSR) score [17] of the bi-cluster $B_k$ is

$$H_k = \frac{\sum_{i \in S_k} \sum_{j \in F_k} r_{ij}^2}{|S_k||F_k|} \tag{6}$$

A sub-matrix $B_k$ is called a $\delta$-bi-cluster if $H_k \leq \delta$ for some $\delta \geq 0$. When calculating the MSR of a single row or single column in data matrix $A$, (6) will be converted to the following expressions:

$$H_{ik}^{(r)} = \frac{1}{|F_k|} \sum_{j \in F_k} r_{ij}^2 \tag{7}$$

$$H_{jk}^{(c)} = \frac{1}{|S_k|} \sum_{i \in S_k} r_{ij}^2 \tag{8}$$

Bi-clusters can thus be seen as sub-matrices of a matrix representing features of elements. It should be noted that bi-clusters need not to be exclusive nor exhaustive [17]. The well-known CC algorithm obtains an optimum bi-cluster (get as large a $\delta$-bi-cluster as possible) each time by adding and deleting some rows or columns in the original data matrix to reduce the MSR of the whole matrix. The bi-clustering result produced each time is shielded by random numbers. However, the random numbers will result in the phenomenon of interference of random numbers, which in turn impacts the discovery of high quality bi-clusters [3, 5]. Eventually, by the action of random numbers, there would be some elements not satisfying the condition of bi-clustering mistakenly clustered; in addition, the overlapping bi-clusters will also be ignored.

An example of the random number interference is shown in Table I. Assume that 0 and 90 in the shadowed entries are clustered in the previous iteration, and replaced by the random numbers 6 and 9 in brackets, and this makes the sub-matrix (such as $B_k = (S_k, F_k)$, $S_k = \{2,4,5\}$, $F_k = \{1,2,4\}$) which is not a bi-cluster, satisfy the condition of the bi-cluster. This shows the unreasonable aspect of the algorithm.

Table I: An example of the random number interference

|   | Col 1 | Col 2 | Col 3 | Col 4 | … … | Col m |
|---|---|---|---|---|---|---|
| Row 1 | Data | Data | Data | Data | … … | Data |
| Row 2 | 1 | 2 | 0 | 3 | … … | Data |
| Row 3 | Data | Data | Data | Data | … … | Data |
| Row 4 | 4 | 5 | 10 | 0 (6) | … … | Data |
| Row 5 | 7 | 8 | 15 | 90 (9) | … … | Data |
| : | : | : | : | : | … … | Data |
| Row n | Data | Data | Data | Data | … … | Data |

### III. COMPLEX SUB-MATRIX SHIELDING MODEL

In connection with this issue above, this paper presents a new complex sub-matrix shielding model and the solutions are figured out. The proposed approach focuses on improving the iteration in the CC algorithm, which must use random numbers to replace the bi-clustering results. However, the new complex sub-matrix shielding can help the algorithm complete the bi-clustering and avoid the interference of the random numbers in the greedy strategy.

#### A. Construction of a shielding complex sub-matrix

Suppose that $B_k = (S_k, F_k)$ is a bi-cluster of the kth bi-clustering searching. To find a (k+1)th new bi-cluster, the first k bi-clusters that have been obtained should be shielded. When searching the (k+1)th bi-cluster based on the first k shielded bi-clusters, on the one hand, it is desired that the first k bi-clusters be temporarily ignored (to find a new bi-cluster); on the other hand, it is desired the (k+1)th bi-cluster should contain the elements that have been clustered in the first k shielded bi-clusters and satisfy the prespecified condition of bi-lustering. To meet the above requirements, a shielding complex sub-matrix is built as

$$\bar{B}_k = \{1 + \varphi\rho[\text{Imag}(B_k) \neq 0]1j\} * B_k + \varphi\rho(B_k)1j \quad (9)$$

where $1j$ is the imaginary unit, $\varphi (\varphi \geq 1)$ is a shielding factor whose role is to adjust the MSR increments of the row (or column) to be shielded. The MSRs increase with the increase of the shielding factor, but usually we just need them to exceed a previously set threshold value. $*$ denotes the Schur product (Hadamard product) [19] of matrices, and $\rho(\bullet)$ is the unit pulse response function. It is necessary to point out that in the process of the shielding, the action of $\varphi\rho[\text{Imag}(B_k) \neq 0]1j$ in (9) is to avoid being shielded a second time, and that of the latter part $\varphi\rho(B_k)1j$ of (9) is to avoid unsuccessful shielding when encountering 0-value data.

#### B. Implementation

By using the shielding complex sub-matrix $\bar{B}_k$, a series of more satisfactory bi-clusters will be discovered. The specific process is as follows:

First, the first bi-cluster $B_1 = (S_1, F_1)$ is found with the CC algorithm and shielded by using the proposed approach. When searching the kth new bi-cluster $B_k (k = 2, 3, \cdots, K)$ the following two expressions are employed to calculate the MSR of a single row or single column in data matrix $A$ and decide whether to delete it or not.

$$H'^{(r)}_{ik} = \frac{1}{|F_k|}\left|\sum_{j \in F_k} r_{ij}^2\right| \quad (10)$$

$$H'^{(c)}_{jk} = \frac{1}{|S_k|}\left|\sum_{i \in S_k} r_{ij}^2\right| \quad (11)$$

It is easy to see that $H'^{(r)}_{ik} \geq H^{(r)}_{ik}$ and $H'^{(c)}_{jk} \geq H^{(c)}_{jk}$, due to the effect of the shielding complex sub-matrix. When the ith row (or jth column) contains elements in the first k-1 bi-clusters, then

$$H'^{(r)}_{ik} \geq H^{(r)}_{ik}, \quad H'^{(c)}_{jk} \geq H^{(c)}_{jk} \quad (12)$$

Meanwhile, with the aid of the shielding factor $\varphi$, the first k-1 bi-clusters will be ignored in the searching of the kth bi-cluster. In order to make the bi-clusters shielded (obtained) be deleted fast, a new parameter $\alpha$ is introduced. If the MSRs of the rows and columns satisfy the following equations, they will be deleted collectively.

$$H'^{(r)}_{ik} = \frac{1}{|F_k|}\left|\sum_{j \in F_k} r_{ij}^2\right| > \alpha H_k \quad (13)$$

$$H'^{(c)}_{jk} = \frac{1}{|S_k|}\left|\sum_{i \in S_k} r_{ij}^2\right| > \alpha H_k \quad (14)$$

After finishing the shielding, a new (kth) bi-cluster is obtained. To make the new bi-cluster contain all the elements in the dataset that satisfy the prespecified condition, what we need to do next is to add the rows and columns that satisfy the prespecified condition. To avoid the disturbance coming from the shielded data, the rows and columns can be added without violating the requirements, i.e.,

$$H'^{(r)}_{ik} = \frac{1}{|F_k|}\sum_{j \in F_k}\left[\text{real}(r_{ij})\right]^2 \leq H_k \quad (15)$$

$$H'^{(c)}_{jk} = \frac{1}{|S_k|}\sum_{i \in S_k}\left[\text{real}(r_{ij})\right]^2 \leq H_k \quad (16)$$

In the processes of deletion and addition of the rows and columns, the MSR is constantly recalculated and improved until the new optimal bi-cluster is obtained. Obviously, the proposed method can avoid the phenomenon of random interference which impacts the discovery of high quality bi-clusters, and can discover overlapping bi-clusters.

### IV. EXPERIMENTAL STUDIES

The following experiments are designed to test the performance of the proposed approach, where a well-known gene expression dataset yeast (http://arep.med.harvard.edu/biclustering/yeast.matrix) which is one of the most commonly used datasets [3] in bi-clustering is used. The methods try to discover 50 bi-clusters (co-expressed) with the MSR score not larger than 300. The MSR score and the sizes of the bi-clusters (co-expressed genes) which are commonly used to estimate (and validate) the performance of the bi-clustering algorithms is used in the experiments.

For each dataset the algorithms are repeated 10 times and the means and the standard deviations of the experimental results are recorded. The experimental results are plotted in Figs. 1 to 2. It is clear that the proposed algorithm is effective in discovering the quality of bi-clusters with low MSR scores. In addition, the bi-clusters obtained by the proposed method also have larger sizes than discovered by CC method. Thus, the experimental results are in agreement with the theoretical analysis, and compared with the well-known CC method, the proposed method exhibits visible advantages.

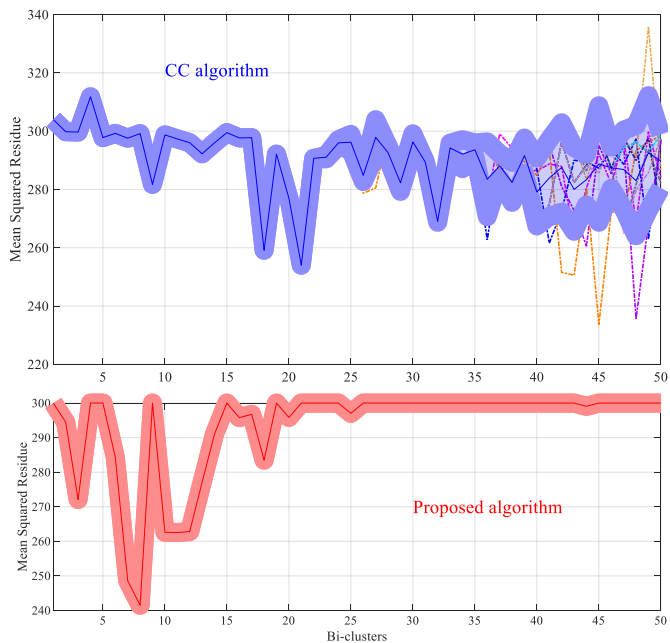

Fig. 1. Results of the MSR scores.

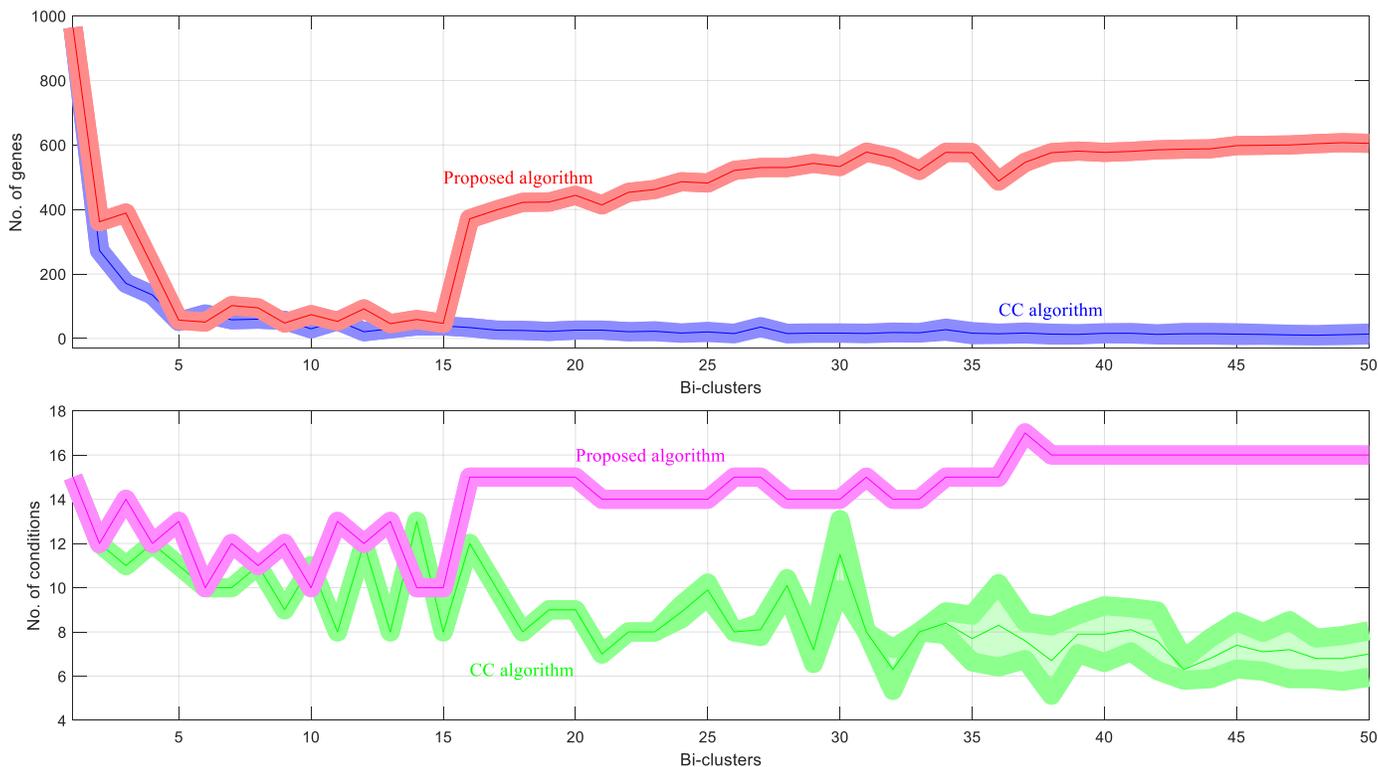

Fig. 2. Results of the Sizes of the co-expressed genes.

## V. CONCLUSIONS

In this research, we designed a novel enhancement adaptive bi-clustering algorithm. During the design process, a shielding complex sub-matrix is built, in which two signals are set to detect the characteristics of dataset and adaptively improve the bi-clusters. We conduct theoretical analysis and offer a comprehensive suite of experiments. Both the theoretical and experimental results are presented to verify the validity of the proposed method. Experiments results show that the proposed algorithm outperforms the CC algorithm in finding the bi-clusters. On the one hand the proposed algorithm can discover overlapping bi-clusters and the MSR is reduced and the sizes of the bi-clusters are increased at the same time. On the other hand, the proposed algorithm is very stable. To the best of our knowledge, this research scheme is first proposed which steadily improves the performance of the bi-clustering. Our result opens a specific way for bi-clustering research, and suggests a far-reaching question for further research: How to discover multiple bi-clusters during a search process? This may open up a new direction of future research pursuits.

REFERENCES

[1] H. Abe, and H. Yadohisa, Orthogonal nonnegative matrix tri-factorization based on Tweedie distributions, Advances in Data Analysis and Classification, 2019, 13(4) 825–853.
[2] Hu H, Wang H, Bai Y. Determination of endometrial carcinoma with gene expression based on optimized Elman neural network, Applied Mathematics and Computation, 341 (2019) 204–214.
[3] Y. Cheng, G. M. Church, Biclustering of expression data, in: Conf. Intelligent Systems for Molecular Biology (ISM), (2000) 93–103.
[4] J. Yang, H. Wang, W. Wang, Enhanced biclustering on expression data, in: Third IEEE Symposium on Bioinformatics and Bioengineering, Bethesda, MD, USA, USA, (2003) 321–327.
[5] J. Yang, H. X. Wang, W. Wang, An improved biclustering method for analyzing gene expression profiles, International journal on artificial intelligence tools, 14 (5) (2005) 771–789.
[6] L. Lazzeroni, A. Owen, Plaid models for gene expression data, Statistica Sinica, 12 (1) (2000) 61–86.
[7] A. Ben-Dor, B. Chor, R. Karp, Discovering local structure in gene expression data: the order-preserving submatrix problem, Journal of Computational Biology, 10 (3) (2003) 373–384.
[8] S. Bergmann, J. Ihmels, N. Barkai, Iterative signature algorithm for the analysis of large-scale gene expression data, Physical review. E, Statistical, nonlinear, and soft matter physics, 67 (1) (2003).
[9] Y. Kluger, R. Basri, T. Chang, Spectral biclustering of microarray data: coclustering genes and conditions, Genome research, 13 (4) (2003) 703–716.
[10] T. M. Murali, S. Kasif, Extracting conserved gene expression motifs from gene expression data, Pacific Symp. Biocomputing, (2003) 77–88.
[11] A. Prelic, S.Bleuler, P. Zimmermann, A systematic comparison and evaluation of biclustering methods for gene expression data, Bioinformatics, 22 (9) (2006) 1122–1129.
[12] J. Gu, J. S. Liu, Bayesian biclustering of gene expression data, BMC Genomics, 1 (2007) 25–28.
[13] C. Huttenhower, K.T. Mutungu, and N. Indik, Detailing regulatory networks through large scale data integration, Bioinformatics, 25 (24) (2009) 3267–3274.
[14] D. Bozdag, J. D. Parvin, U.V. Catalyurek, A biclustering method to discover co-regulated genes using diverse gene expression datasets, in Proc. Proceedings of the 1st International Conference on Bioinformatics and Computational Biology, (2009) 151–163.
[15] G. Li, Q. Ma, H. Tang, QUBIC: a qualitative biclustering algorithm for analyses of gene expression data, Nucleic Acids Research, 37 (15) (2009) e101.
[16] S. Hochreiter, U. Bodenhofer, and M. Heusel, FABIA: factor analysis for bicluster acquisition, Bioinformatics, 26 (12) (2010) 1520–1527.
[17] F. Xhafa, S. Caballe, L. Barolli, Using bi-clustering algorithm for analyzing online users activity in a virtual campus, in: International Conference on Intelligent NETWORKING and Collaborative Systems, Thessaloniki, Greece, (2011) 214–221.
[18] N. Fan, N. Boyko, P. M, Pardalos recent advances of data biclustering with application in computational neuroscience, Computational Neuroscience. Springer, New York, USA, (2010) 85–112.
[19] K. J. Xu, W. Pedrycz, Z. W. Li, W. K. Nie, High-accuracy signal subspace separation algorithm based on gaussian kernel, IEEE Transactions on Industrial Electronics, 66 (1) (2019) 491–499.